\documentclass[runningheads]{llncs}

\usepackage{eccv}

\usepackage{eccvabbrv}
\usepackage{balance}
\def\eqref#1{(\cref{#1})}
\usepackage[utf8]{inputenc} 
\usepackage[T1]{fontenc}    
\usepackage{url}            
\usepackage{booktabs}       
\usepackage{amsfonts}       
\usepackage{nicefrac}       
\usepackage{microtype}      
\usepackage{xcolor}         
\usepackage{microtype}
\usepackage{tablefootnote}
\usepackage{subcaption,graphicx}
\usepackage{graphicx}
\usepackage{booktabs}
\usepackage{booktabs} 
\usepackage{float}
\usepackage{array,multirow}

\usepackage[ruled, vlined, linesnumbered]{algorithm2e}
\usepackage{amsmath}
\usepackage{amssymb}
\usepackage{mathtools}
\usepackage{tabularx}
\usepackage[export]{adjustbox}
\usepackage{wrapfig}
\usepackage{amssymb}
\usepackage{pifont}
\newcommand{\cmark}{\ding{51}}%
\newcommand{\xmark}{\ding{55}}%

\newcommand{\mbs}{{\mathbf{S}}}
\newcommand{\mbc}{{\mathbf{C}}}
\newcommand{\mby}{{\mathbf{Y}}}
\newcommand{\mbq}{{\mathbf{Q}}}
\newcommand{\mbk}{{\mathbf{K}}}
\newcommand{\mbv}{{\mathbf{V}}} 
\newcommand{\hii}{{\mathbf{ h}_{\text{img}}}}
\newcommand{\hht}{{\mathbf{h}_{\text{txt}}}}
\newcommand{\htt}{{\mathbf{\tilde h}_{\text{txt}}}}
\newcommand{\mbty}{{\mathbf{\tilde Y}}}
\newcommand{\asty}{{\mathbf{Y}^*}}
\newcommand{\gty}{\mbf{Y}^{\text{gt}}}
\newcommand{\mc}{\mathcal}
\newcommand{\mbf}{\mathbf}
\newcommand{\bs}{\mathbf}
\usepackage[accsupp]{axessibility}  
\usepackage{wrapfig}

\usepackage[pagebackref,breaklinks,colorlinks]{hyperref}

\usepackage{orcidlink}

\begin{document}

\title{{OTSeg: {Multi-prompt Sinkhorn Attention} for Zero-Shot Semantic Segmentation}}

\author{Kwanyoung Kim$^{*1}$\orcidlink{0000-0001-7508-7145}, Yujin Oh$^{*2}$\orcidlink{0000-0003-4319-8435}, and Jong Chul Ye$^{1}$\orcidlink{0000-0001-9763-9609}, Fellow, IEEE
}

\titlerunning{OTSeg}


\authorrunning{K. Kim, Y. Oh and J. C. Ye}

\institute{Graduate School of AI, Korea Advanced Institute of Science and Technology (KAIST), South Korea \and
Department of Radiology, Massachusetts General Hospital (MGH) and
Harvard Medical School, Boston, MA, USA\\
{$^{*}$Equal contribution.}
}

\maketitle

\begin{abstract}
{
The recent success of CLIP has demonstrated promising results in zero-shot semantic segmentation by transferring muiltimodal knowledge to pixel-level classification. However, leveraging pre-trained CLIP knowledge to closely align text embeddings with pixel embeddings still has limitations in existing approaches. To address this issue, we propose OTSeg, a novel multimodal attention mechanism aimed at enhancing the potential of multiple text prompts for matching associated pixel embeddings. We first propose Multi-Prompts Sinkhorn (MPS) based on the Optimal Transport (OT) algorithm, which leads multiple text prompts to selectively focus on various semantic features within image pixels. Moreover, inspired by the success of Sinkformers in unimodal settings, we introduce the extension of MPS, called  Multi-Prompts Sinkhorn Attention (MPSA), which effectively replaces cross-attention mechanisms within Transformer framework in multimodal settings. Through extensive experiments, we demonstrate that OTSeg achieves state-of-the-art (SOTA) performance with significant gains on Zero-Shot Semantic Segmentation (ZS3) tasks across three benchmark datasets. We release our source code at \url{https://github.com/cubeyoung/OTSeg}.
}
\end{abstract}


\begin{figure*}[!t]
\centering
\includegraphics[width = 0.9\linewidth]{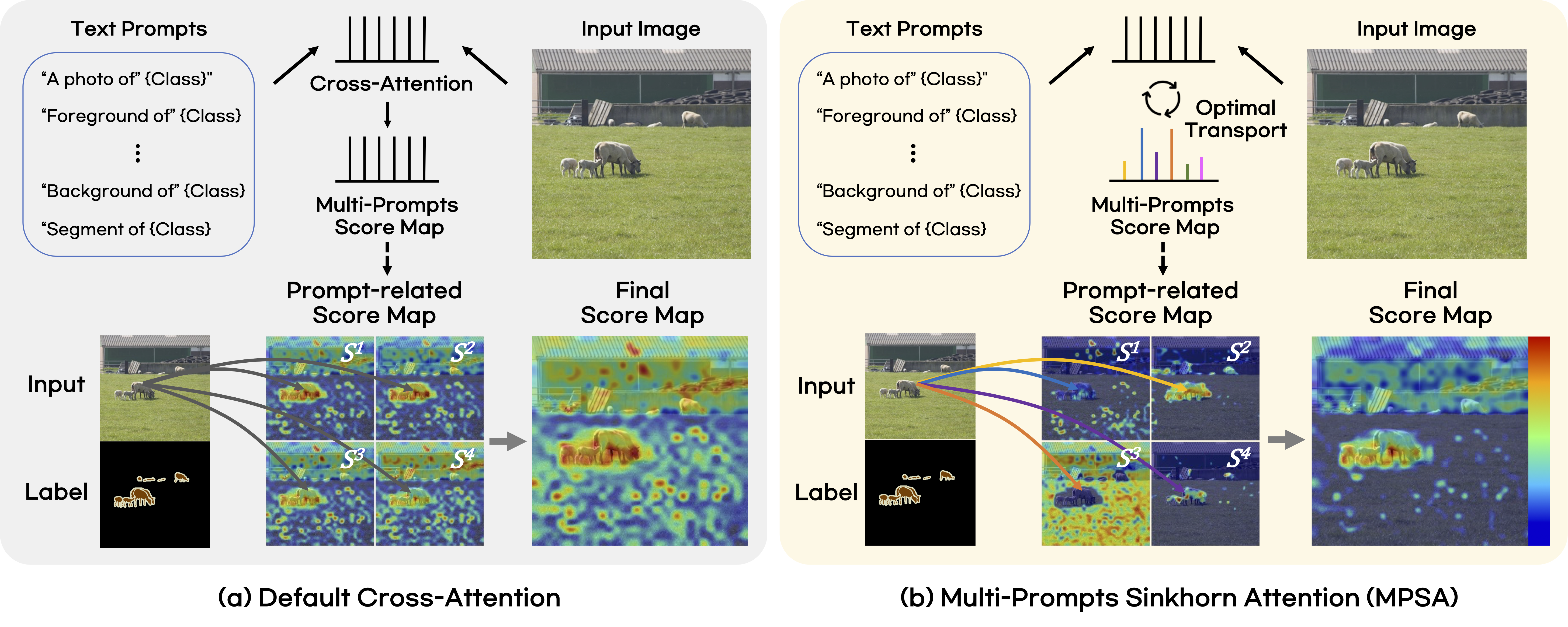}
\vspace{-1em}
\caption{Visualization of proposed Multi-Prompts Sinkhorn Atttention (MPSA) for text-driven semantic segmentation. (a) Without MPSA, {all the text prompt-related score maps ${S}^i$ are cohered}. (b) With MPSA, each ${S}^i$ selectively focuses on different semantic attributes, resulting the final score map effectively attends to the target object.}

\label{fig_intro}
\vskip -0.2in
\end{figure*}


\section{Introduction}
\label{submission}

{Transformer's attention mechanism has demonstrated its remarkable performance across various tasks, establishing itself as a universal backbone structure in foundational models \cite{radford2021clip, caron2021emerging, devlin2018bert, touvron2023llama}. However, recent advancements 
cast doubt on the optimality of traditional transformer structures that rely solely on consecutive self-attention layers. Among widespread  efforts to enhance transformer performance  \cite{kim2020t, nguyen2022improving, nguyen2022improving2}, one notable endeavor is  Sinkformer \cite{sander2022sinkformers}, where the SoftMax normalization of self-attention mechanism within transformer is simply replaced by the optimal transport (OT) algorithm, yielding a doubly stochastic attention that improves model performance in each vision and linguistic task. 
}

{While Sinkformer is exclusively designed for unimodal settings, we are interested in its more appropriate applicability for multimodal alignment, particularly in the context of text-driven semantic segmentation tasks. In text-driven semantic segmentation, zero-shot semantic segmentation (ZS3) \cite{bucher2019zero} represents a label-efficient approach. A key factor contributing to the recent advancements in ZS3 solutions \cite{ding2022zegformer, xu2021zsseg, zhou2022maskclip, rao2022denseclip, zhou2023zegclip} is application of pre-trained Vision-Language Model (VLM), such as CLIP\cite{radford2021clip}. 
However, care should be taken when transferring VLM, as the pre-trained knowledge is not optimized for pixel-wise dense alignment, since VLM has been trained with a variety of image-text pairs through contrastive learning. One naive solution can be fine-tuning the VLM tailored for desired pixel-level prediction by leveraging an ensemble technique driven by multiple text prompts \cite{pmlr-v202-allingham23a}. 
However, as shown in \cref{fig_intro}(a), this naive approach is still problematic because the introduced multiple text prompts and image pixels are passively aligned, which leads each text prompt-driven pixel-level prediction to be cohered to each other.}


{To address the limited potential of multiple text prompts, we introduce a pixel-text alignment operator based on the OT algorithm, namely, Multi-Prompts Sinkhorn (MPS). Furthermore, as like the aforementioned Sinkformer improved the traditional transformer by introducing doubly stochastic self-attention, we have discovered that the MPS algorithm can serve as an optimal replacement for the SoftMax normalization of cross-attention mechanism within multimodal transformer. This leads to the development of a novel Multi-Prompts Sinkhorn Attention (MPSA) module, which composes our proposed OTSeg.
As shown in \cref{fig_intro}(b), we empirically find that our OTSeg enhances the diversity of text prompt-driven pixel-level predictions by selectively focusing on various semantic features. Consequently, the optimally ensembled final prediction effectively focuses on the target object, leading to the achievement of the state-of-the-art (SOTA) performance in ZS3 tasks.
Our contributions can be summarized as:}
\begin{itemize}

\item {We introduce a novel OTSeg that yields diverse pixel-level predictions driven by multiple text prompts, resulting in improved multimodal alignment.}
\item {Through extensive experiments on three benchmark datasets, OTSeg demonstrates its superiority on ZS3 tasks by achieving SOTA performance.}

\end{itemize}

\section{Related Work}


\subsection{Zero-shot {\& Open-vocabulary} Semantic Segmentation}
Semantic segmentation is a core computer vision task to densely analyze visual context. Recent success of CLIP \cite{radford2021clip} accelerates the advancement of language-driven semantic segmentation by utilizing pre-{trained knowledge of VLM }
\cite{li2022language, xu2022groupvit, liang2022open}.
However, since this dense prediction task requires a labor-intensive pixel-level annotation, {there arise a label-imbalance issue,} \textit{i.e.,} not all the categories are annotated in the training dataset. 
Zero-shot semantic segmentation (ZS3) solves this label-imbalance problem by generalizing labeled (seen) class knowledge to predict new (unseen) class information \cite{bucher2019zero}. 
MaskCLIP+ \cite{zhou2023zegclip} introduces a ZS3 method by simply extracting the text-driven visual features from the CLIP image encoder. 
ZegCLIP \cite{zhou2023zegclip} successfully bridges the performance gap between the seen and unseen classes by adapting a visual prompt tuning technique instead of fine-tuning the frozen CLIP image encoder. 
{Recently,} FreeSeg \cite{qin2023freeseg} and 
MVP-SEG+ \cite{guo2023mvp} {introduce text prompt-driven method for realizing open-vocabulary segmentation. In specific, MVP-SEG+ employs orthogonal constraint loss (OCL) to each prompt} 
to exploit CLIP features on different object parts. 

ZS3 can be performed by either inductive or transductive settings. Compared to inductive ZS3 where class names and pixel-level annotations of unseen classes are both unavailable during training \cite{ding2022zegformer}, a newly introduced transductive setting boosts the ZS3 performance by utilizing unseen class names and self-generated pseudo labels guided by the model itself during training \cite{gu2020cagnet, xu2021zsseg, pastore2021strict, zhou2022maskclip, zhou2023zegclip}. {Open-vocabulary settings simply extend the concept of inductive ZS3 settings applied for cross-domain datasets. In this study, we apply our proposed OTSeg for both the inductive and transductive settings, and we further demonstrate our OTSeg performance on various {cross-dataset settings.}

\subsection{Optimal Transport}
Optimal transport (OT) is a general mathematical framework to evaluate correspondences between two distributions. 
Thanks to the luminous property of distribution matching, the optimal transport has received great attention and proven its generalization capability in various computer vision tasks, such as domain adaptation~\cite{flamary2016optimal}, semantic correspondence problem~\cite{liu2020semantic}, graph matching~\cite{xu2019scalable,xu2019gromov}, and cross-domain alignment~\cite{chen2020graph}, etc.  
Among various methods, Sinkhorn algorithm can efficiently solve the OT problem through entropy-regularization~\cite{cuturi2013sinkhorn}, and it can be directly applied to deep learning frameworks thanks to the extension of Envelop Theorem~\cite{peyre2019computational}. 
{In the context of computer vision tasks,} prompt learning with optimal transport (PLOT) \cite{chen2022prompt} optimizes the OT distance to align visual and text features by the Sinkhorn  given trainable multiple text prompts for few-shot image-level prediction tasks. 
{The most related work to ours is Sinkformer\cite{sander2022sinkformers}, where SoftMax layers within self-attention transformer is replaced by Sinkhorn algorithm, resulting in enhanced accuracy in each vision and natural language processing task. {The Sinkformer has inspired us to propose the Sinkhorn algorithm as an ideal fit for multimodal alignment, leading us to apply OT to further boost the performance of the cross-attention mechanism for ZS3 tasks.}}


\section{Preliminary}
\subsection{Optimal Transport Problem and Sinkhorn}
\label{sec:sinkhorn}
Optimal transport aims to minimize the transport distance 
between two probability distributions. In this paper, we only consider discrete distribution which is closely related to our framework. We assume the feature vector $F,G$ are defined as $F = \{\boldsymbol{f}_i \}_{i=1}^M$ and $G = \{\boldsymbol{g}_j\}_{j=1}^N$ and discrete empirical distributions $\mathbf{u}$ and $\mathbf{v}$ that are defined on probability space $\mathcal{F}, \mathcal{G} \in \Omega$, respectively, as follows:
\begin{align}
\mathbf{u} = \sum^{M}_{i=1} \mu_{i} \delta_{\boldsymbol{f}_i}, \quad \mathbf{v} = \sum^{N}_{j=1} \nu_{j} \delta_{\boldsymbol{g}_j}, 
\end{align}
where $\delta_{\boldsymbol{f}}$ and $\delta_{\boldsymbol{g}}$ denote Dirac functions centered on $\boldsymbol{f}$ and $\boldsymbol{g}$, respectively, $M$ and $N$ denote the dimension of the empirical distribution. 
The weight vectors $\boldsymbol{\mu} = \{\mu_i\}^M_{i=1}$ and $\boldsymbol{\nu} = \{\nu_j\}^{N}_{j=1}$  belong to the $M$ and $N$-dimensional simplex, respectively, \textit{i.e.}, $\sum^{M}_{i=1} \mu_i = 1$, $\sum^{N}_{j=1} \nu_j = 1$. The discrete optimal transport problem  can be then formulated as: 
\begin{align}
\bs{T}^{\ast} = \underset{\bs{T}\in \mathbb{R}^{MXN}}{\arg{\min}} \sum^{M}_{i=1}\sum^{N}_{j=1}\bs{T}_{ij} \bs{C}_{ij} \nonumber \\ \textrm{s.t.} \quad \bs{T}\bs{1}^{N} = \boldsymbol{\mu}, \quad \bs{T}^{\top}\bs{1}^{M} = \boldsymbol{\nu} .
\label{DOT}
\end{align}
Here,
$\bs{T}^{\ast}$ is called the optimal transport plan, which is learned to minimize 
the total distance between the two probability vectors. $\bs{C}$ is the cost matrix which represents the distance between $\boldsymbol{f}_i$ and $\boldsymbol{g}_j$, \textit{e.g.,} the cosine distance $\bs{C}_{ij}$ = 1 - $\frac{\bs{f}_i\bs{g}^{\top}_j}{\|\bs{f}_i\|_2 \|\bs{g}_j\|_2}$, and $\bs{1}^{M}$ refers to the $M$-dimensional vector with ones. 

However, solving the problem~\cref{DOT} costs $O(n^3\log n)$-complexity ($n$ proportional to $M$ and $N$), which is time-consuming. 
This issue can be efficiently solved by the entropy-regularization of the objective through the Sinkhorn-Knopp (or simply Sinkhorn) algorithm~\cite{cuturi2013sinkhorn}. In
Sinkhorn algorithm, the optimization problem is reformulated as:
\begin{align}
\bs{T}^{\ast} = \underset{\bs{T}\in \mathbb{R}^{MXN}}{\arg{\min}} \sum^{M}_{i=1}\sum^{N}_{j=1}\bs{T}_{ij}\bs{C}_{ij} - \epsilon H(\bs{T}) \nonumber \\ \textrm{s.t.} \quad \bs{T}\bs{1}^{N} = \boldsymbol{\mu}, \quad \bs{T}^{\top}\bs{1}^{M} = \boldsymbol{\nu} .
\label{Sinkhorn}
\end{align}
where $H(\bs{T})$ = $\sum_{ij} \bs{T}_{ij} \log \bs{T}_{ij}$ and $\epsilon > 0$ is the regularization parameter.
For the problem \cref{Sinkhorn}, 
we have an optimization solution when $t \rightarrow \infty$ as follow: 
\begin{align}
 \bs{T}^{\ast} = \text{diag}(\bs{a}^{t})\exp(-\bold{C}/\epsilon)\text{diag}(\bs{b}^{t})
\label{Sinkhorn2}
\end{align}
where $t$ is the iteration and $\bs{a}^t = \boldsymbol{\mu}/ \exp(-\bold{C}/\epsilon)\bs{b}^{t-1}$ and $\bs{b}^{t} = \boldsymbol{\nu}/\exp(-\bold{C}/\epsilon)\bs{a}^{t}$, with the initialization on $\bs{b}^{0}=\bs{1}$. To stabilize the iterative computations, we adopt the log scaling version of Sinkhorn optimization~\cite{schmitzer2019stabilized}. 


\begin{figure*}[!t]
\centering
\includegraphics[width = 0.9\linewidth]{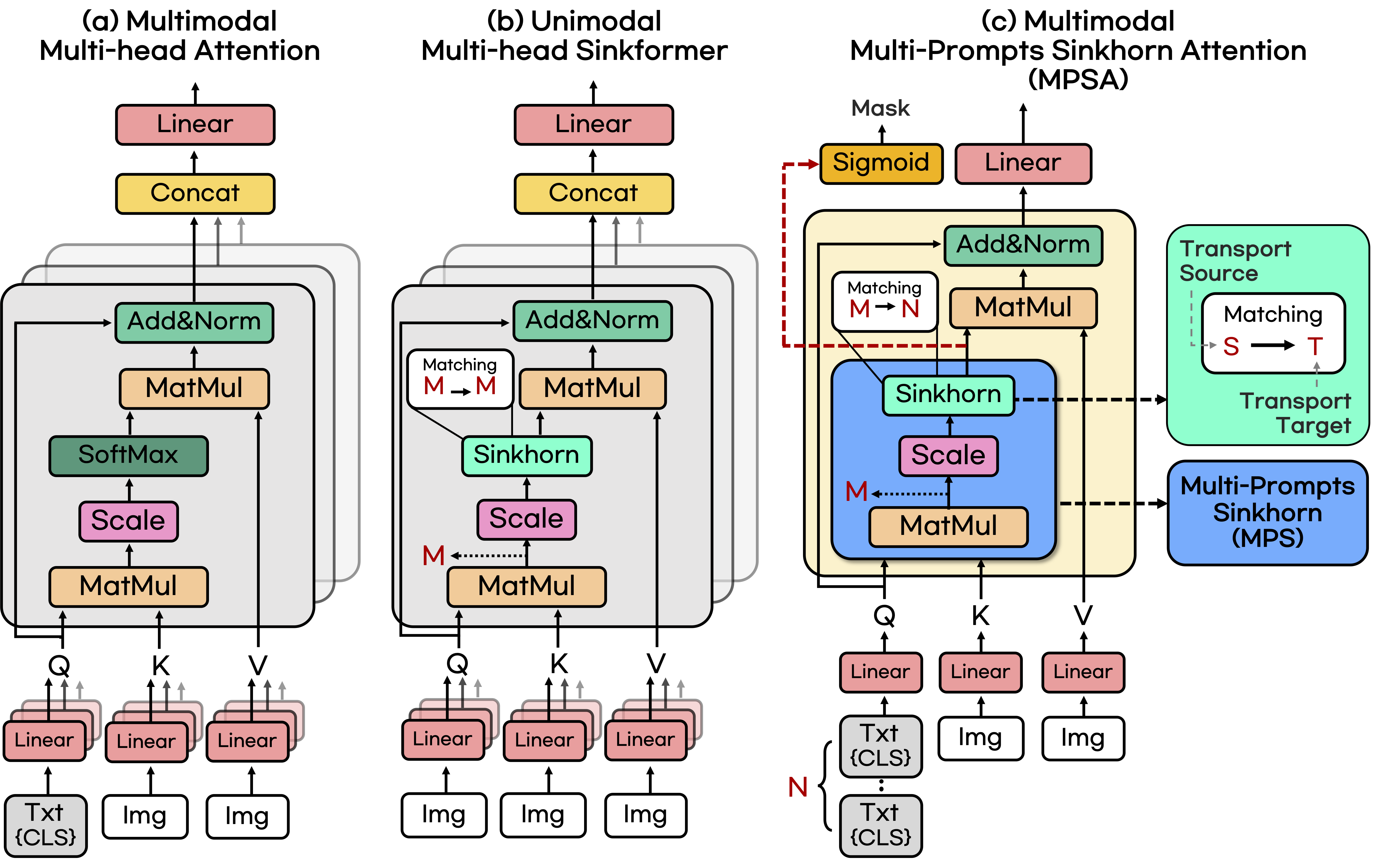}
\vspace{-0.5em}
\caption{{Comparison of attention mechanism variants. (a) Cross-attention mechanism for multimodal settings. (b) Sinkformer self-attention mechanism for unimodal settings. (c) Our proposed Muti-Prompt Sinkhorn Attention (MPSA) for multimodal settings, which aims to optimally transport image pixel (M) to multiple text prompts (N).}}
\label{fig_main}
\vskip -0.2in
\end{figure*}

\subsection{Self-Attention and Sinkformer}
The transformer model employs a self-attention mechanism when given a sequence of length $\mathbf{X}$ =[$x_1,x_2, \cdots, x_n$] embedded in $d$ dimensions:
\begin{align}
\text{Self-Attention}(\mbq,\mbk,\mbv) = \texttt{SoftMax}\left(\frac{\mbq \mbk^{\top}}{\sqrt{d}}\right)\mbv \\ \label{self-atten}
\text{where}\quad \mathbf{Q} = \phi_q(\mathbf{X}), \; \mathbf{K} = \phi_k(\mathbf{X}), \; \mathbf{V} = \phi_v(\mathbf{X}),
\end{align}
where $\mbq, \mbk$ $\in \mathbb{R}^{M \times d}$, and $\mbv \in \mathbb{R}^{d \times d} $ denotes query, key and value matrices, respectively. $\phi_q, \phi_k$ and $\phi_v$ represent the linear layer of each matrix.
In Eq.(\ref{self-atten}), the attention matrix is normalized using the \texttt{SoftMax} operator. In Sinkformer, it is demonstrated that the initial iteration in the Sinkhorn algorithm is identical to the softmax operation. This observation leds to the replacement of \texttt{SoftMax} with the Sinkhorn algorithm in self-attention to ensure double stochasticity. For simplicity, Sinkhorn's algorithm is seamlessly integrated into the self-attention modules as follow:
\begin{align}
\text{Sinkformer Attention}(\mbq,\mbk,\mbv) = \texttt{Sinkhorn}\left(\frac{\mbc}{\sqrt{d}}\right)\mbv, \; \mbc = \boldsymbol{1} - \mbq \mbk^{\top}
\end{align}
where \texttt{Sinkhorn}($\cdot$) is the Sinkhorn operator, which reduces the cost $\mbc$ by utilizing \cref{Sinkhorn2}. The scale factor $\epsilon$ is replaced by $\sqrt{d}$. Despite the extension, Sinkformer is still limited in unimodal settings such as the self-attention within an image modality {(M pixels$\rightarrow$M pixels),} as illustrated in \cref{fig_main}(b). 


\begin{figure*}[!t]
\vskip 0.1in
\begin{center}
\includegraphics[width=0.9\linewidth]{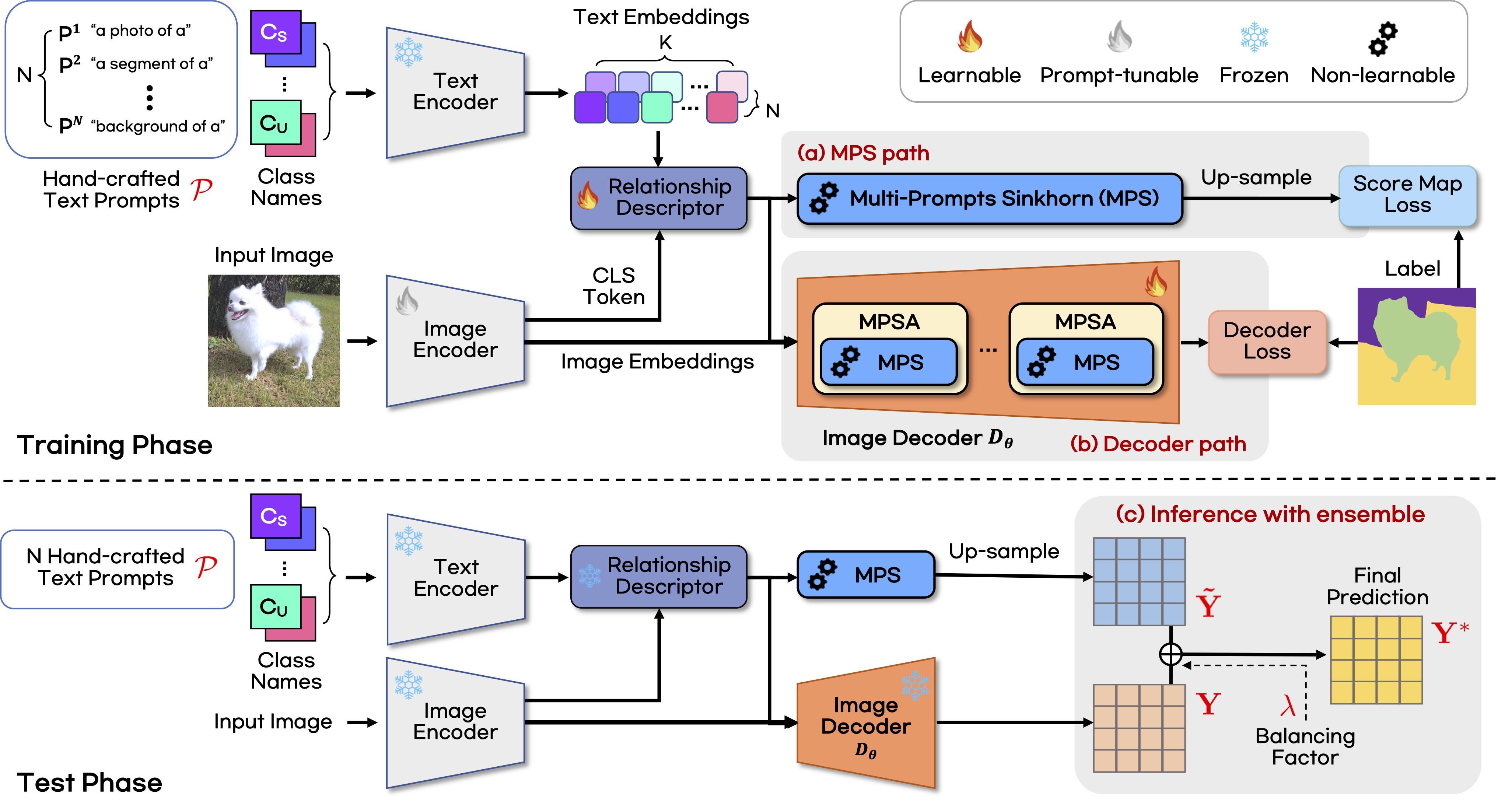}
\caption{
Overview of OTSeg for zero-shot semantic segmentation. (a) MPS path refines the score map using the \texttt{MPS} algorithm.
(b) Decoder path involves the decoder output, which integrates the Multi-Prompts Sinkhorn Attention (MPSA) predictions.
(c) During inference, OTSeg ensembles predictions from both paths with a balancing factor $\lambda$. }

\label{fig_main0}
\end{center}
\vspace{-2em}
\end{figure*}

\section{Methods}
\label{methods}
In this section, we present a method for performing zero-shot segmentation tasks using our proposed  {OTSeg framework for multimodal setting {(M pixels$\rightarrow$N text prompts)}, as illustrated in \cref{fig_main}(c).} 
To clarify, we define several notations within our OTSeg framework: a pair of frozen CLIP text encoder and tunable image encoder, the relationship descriptor, multiple hand-crafted text prompts $\mathcal{P}$, and trainable decoder $D_{\theta}$. Furthermore, we propose three fundamental components of OTSeg: (a) Multi-Prompts Sinkhron (MPS), (b) Multi-{Prompts} Sinkhorn Attention (MPSA) and (c) Inference with ensemble.x
In \cref{task}, we introduce the concept of multiple prompts-guided text embeddings and explain how they are processed. 
In \cref{MPS}, we provide a detailed account of our OTSeg, including aforementioned three key components. Lastly, in \cref{sec:loss}, we describe the training procedure for our method and introduce the associated loss functions.




\subsection{Multiple Prompts-guided Text Embeddings}
\label{task}
To effectively transfer CLIP's pre-trained knowledge, we adopt the frozen text encoder with multiple hand-crafted text prompts and a tunable image encoder, as shown in \cref{fig_main0}. The multiple text prompts are created as a set of $N$ text prompts denoted as $\mathcal{P} = \{{\bs{P}^{i}|^N_{i=1}}\}$. Each text prompt $\bs{P}^i$ can be defined as $\bs{P}^i =[P^i_1,P^i_2, \cdots, P^i_l]$, where $l$ represents the length of the context tokens.
These text prompts are then consistently added in front of $K$ tokenized class names, forming a set denoted as $\mathcal{T} = \{\{\mathcal{P}, \bs{c}^k\}|^K_{k=1}\}$. Note that the same text prompts $\mathcal{P}$ are shared across all class names. 
Here, $\{\bs{c}^k|_{k=1}^K\}$ represents the word embeddings of each class name, drawn from a larger set $\mc{C}$. Then, the set $\mathcal{T}$ is inputted to the frozen CLIP text encoder, resulting in the text embeddings $\hht \in {\mathbb{R}^{KN \times D}}$, where $D$ represents the embedding dimension. We utilize the relationship descriptor following an approach introduced in~\cite{zhou2023zegclip} to yield the refined text embedding $\htt \in {\mathbb{R}^{KN \times D}}$. {The detail of relationship descriptor are deferred in Appendix~\ref{appendix:RD}.}


\subsection{Optimal Transformer for Zero-Shot Segmentation}

\textbf{(a) Multi-Prompts Sinkhorn (MPS)} \label{MPS}
Now, when we input an image through the tunable CLIP image encoder, the model yields pixel embedding $\mathbf{h}_{\text{img}} \in \mathbb{R}^{M \times D}$, where $M = H \times W$ corresponds to the product of the height $H$ and width $W$.
Given the text embeddings and pixel embeddings, the text-pixel aligned score map can be formulated as follow: 
\begin{align}
	\mbs = \htt \hii^{\top},
	\label{eq:score}
\end{align} 
where the superscript $^{\top}$ refers to the transpose operation, both $\htt$ and $\hii$ are $\mathcal{L}_2$ normalized along the embedding dimension, and the score map $\mbs \in \mathbb{R}^{HW \times KN}$ 
undergoes further refinement. 

To transport the distribution of the multiple text prompts to pixel distribution, we first define the total cost matrix $\bs{C}$ in \cref{Sinkhorn} using the text-pixel aligned score map $\bs{S}$ in~\cref{eq:score}. Specifically, we set 
{$\bs{C} := \bs{1 - S}$,} where $\bs{C} \in \mathbb{R}^{HW \times KN}$ denotes the cost matrix.
Given the cost matrix $\bs{C}$, the goal of \texttt{MPS} is to obtain the corresponding optimal transport plan $\bs{T}^{\ast}$ as given in \cref{Sinkhorn2}, {which aims to allocate each of the $M$ image pixels to the $N$ text-prompts, thereby allowing multiple text prompts to be associated with each pixel. Thus, $\bs{T}^{\ast}$ serves as a mapping matrix that maximizes the cosine similarity between multimodal embeddings, as outlined as \cref{algo-ot} in \cref{sec_algo}.} Therefore, we formulate the refined score map through \texttt{MPS} algorithm: 
\begin{align}
&\bs{S}^{\ast} = \texttt{MPS}(\bs{S}) =  \mc{M}(\bs{T}^{*}\odot\bs{S}) \label{eq:mps} \\ 
\text{where} \; &  \; \bs{T}^{*} = \texttt{Sinkhorn}\left(\frac{\mbc}{\epsilon}\right), \; \bs{C} := \bs{1 - S} \; 
\end{align}
where $\mathcal{M}$: $\mathbb{R}^{HW \times KN} \rightarrow \mathbb{R}^{HW \times K}$ is the operation which first reshapes $\mathbb{R}^{HW \times KN} \rightarrow \mathbb{R}^{H W \times K \times N}$ and performs summation of all the score maps along the $N$ dimension, $\epsilon$ denotes the scaling hyper-parameter.  The refined score map $\bs{S}^{\ast} \in \mathbb{R}^{HW \times K}$ by adapting the transport plan $\bs{T}^{\ast}$ can be served as a stand-alone logit for segmentation mask as outlined in \cref{infer} (c). 
Note that the transport plan $\bs{T}^{\ast}$ in~\cref{Sinkhorn2} only contains matrix multiplication and exponential operation, thus \texttt{MPS} algorithm is fully differentiable and the gradients can be back-propagated throughout the entire neural network. \\

\noindent\textbf{(b) {Muti-Prompts Sinkhorn Attention (MPSA) } \label{MPSA} }
In the multimodal settings for text-driven semantic segmentation tasks, we can formulate cross-attention between pixel embeddings and classnames-driven text embeddings {with multiple prompts}:
\begin{align}
&\text{{Cross-Attention}} 
(\mbq,\mbk,\mbv) = \texttt{SoftMax}\left(\frac{\mbq\mbk^{\top}}{\sqrt{d}} \right)\mbv \label{cross}\\ 
&  \text{where} \mbq = \phi_q(\htt), \; \mbk = \phi_k(\hii), \mbv = \phi_v(\hii),
\end{align}
{where $\mbq \in \mathbb{R}^{KN \times D}$, and $\mbk, \mbv$ $\in \mathbb{R}^{M \times D}$} denotes query, key and value matrices, respectively, and $\phi$ denotes linear projection for each query, key, value. Instead of applying the \texttt{MPS} algorithm solely on the score map, we empirically find that \texttt{MPS} can be extended into the cross-attention mechanism and seamlessly integrated as a plugin module in each decoder layer. 
Similar to the cross-attention mechanism~\cref{cross}, we define our proposed Multi-Prompts Sinkhorn Attention as follows:
\begin{align} 
&\text{{Multi-Prompts Sinkhorn Attention}} 
(\mbq,\mbk,\mbv) = \texttt{MPS}\left(\mbq\mbk^{\top} \right)\mbv. 
\end{align}
In our proposed MPSA, instead of conventional multiple head dimensions, we have $N$ multiple text-prompts dimensions, as shown in \cref{fig_main}(c). {In our decoder, comprising three layers of cross-attention transformer as illustrated in \cref{fig_main0}, each multi-head attention module is replaced by our proposed Multi-Prompts Sinkhorn Attention (MPSA) module.} {In the decoder, a semantic mask is calculated by taking the intermediate product of MPSA, denoted by the \texttt{MPS} operation, followed by the \texttt{Sigmoid} function:}
\begin{align}
\texttt{Mask} &= \texttt{Sigmoid}\left(\texttt{MPS}(\mbq\mbk^{\top})\right). 
\end{align}
where {$\texttt{Mask}$ denotes the semantic mask in the final decoder layer {as shown in \cref{fig_main}(c)}, while $\mbq$ and $\mbk$ refer to the query and key matrices from the preceding layer, respectively.}

Then, we can obtain the final decoder output by applying the up-sampling operator as follow:
\begin{align}
\mby &= \mc{U}(\texttt{Mask}), \; \in \mathbb{R}^{H_I W_I \times K} \label{eq:predict1} 
\end{align}
where $\mby \in \mathbb{R}^{H_IW_I \times K}$ is the final output of decoder which is integrated in our MPSA module, $\mathcal{U}:\mathbb{R}^{HW \times K} \rightarrow \mathbb{R}^{H_I W_I \times K}$ is the up-sampling operator ($HW < H_I W_I$), where $H_I$ and $W_I$ are height and width of the input image, respectively. {When the prediction is calculated from \cref{eq:predict1} using decoder, we refer it as OTSeg.}\\

\noindent\textbf{(c) Inference with ensemble} \label{infer}
Rather than solely relying on the prediction in~\cref{eq:predict1}, we further utilize the refined score map $\bs{S}^{\ast}$ in \cref{eq:mps} to boost the segmentation performance.
For this purpose, $\mathbf{S}^{\ast}$ is up-sampled to match the original image size as follows:
\begin{align}
\mbty = \mathcal{U}(\bs{S}^{*}).
 \label{eq:predict2}
\end{align}
where $\mbty \in \mathbb{R}^{H_IW_I \times K}$ is the prediction of the refined score map.
{In order to synergistically exploit the collective knowledge derived from the learnable decoder in $\mby$ and  information encapsulated in $\mbty$}, we formulate the final segmentation output $\asty$ as follows:
\begin{align}
\asty =  \lambda \cdot \mby  +  (1 - \lambda)\cdot \mbty
 \label{eq:predictfinal}
\end{align}
where balance factor $\lambda \in [0,1]$ denotes the hyper-parameter for controlling balance between $\mby$ and $\mbty$, and set to 0.5 through component analysis in Appendix.
{We refer to this ensembled approach as OTSeg+.} 



\subsection{Loss Function}  \label{sec:loss} 
In this work, we combine two different losses following previous methods as follows:
\begin{align}
\mathcal{L}_{\text{seg}} = \lambda_{\text{fc}} \mathcal{L}_{\text{fc}} + \lambda_{\text{dc}} \mathcal{L}_{\text{dc}}, 	\;  \mathcal{L}_{\text{tot}}(\Theta,\theta) = 
\mathcal{L}_{\text{seg}}(\mby,\gty;\Theta, \theta) + \mathcal{L}_{\text{seg}}(\mbty, \gty;\Theta)
\label{losses}
\end{align}
where $\Theta = [E_{\text{img}},\mathcal{R}_{\psi}]$ contains tunable image encoder $E_{\text{img}}$ and the linear layer $\mathcal{R}_{\psi}$, $\mathcal{L}_{\text{seg}}$ denotes the segmentation loss combining different two losses, $\mathcal{L}_{\text{fc}}$ and $\mathcal{L}_{\text{dc}}$ are the focal loss, and the dice loss, with $\lambda_{\text{fc}}$, and  $\lambda_{\text{dc}}$ as corresponding hyper-parameters, respectively, and
$\gty \in \mathbb{R}^{H_I W_I \times K}$ is the ground-truth label. The details of the loss function are described in \cref{appendix:loss}.

\section{Experiments}

\subsection{Dataset and Evaluation Metric}
\textbf{Dataset}
A primary goal of ZS3 is to segment objects belong to both seen classes $\mc{C}_S$ and unseen classes $\mc{C}_U$, \textit{i.e.,} $\mc{C} = \mc{C}_{S} \cup \mc{C}_U$, where $\mc{C}_S \cap \mc{C}_U = \emptyset$. 
For fair comparison with previous methods~\cite{bucher2019zero,xu2021zsseg,ding2022zegformer,zhou2022maskclip,zhou2023zegclip}, we follow the identical protocol of dividing $\mc{C}_S$ and $\mc{C}_U$ for each dataset. 
To evaluate the effectiveness of our OTSeg, we carry out extensive experiments on three challenging datasets: VOC 2012~\cite{everingham2012pascal}, PASCAL  Context~\cite{mottaghi2014role}, and COCO-Stuff164K~\cite{caesar2018coco}.
The dataset details are described in~\cref{appendix:dataset}.  \\

\noindent\textbf{Evaluation Metric}
We measure the mean of class-wise intersection over union (mIoU) on both seen and unseen classes, indicated as mIoU(S) and mIoU(U), respectively. We adopt the harmonic mean IoU (hIoU) of seen and unseen classes as a primary metric. More details are deferred to~\cref{appendix:hIoU}. 


\subsection{Implementation Details}
We implemented the proposed method using the open-source toolbox MMSegmentation~\cite{mmseg2020}. 
The algorithm was executed on up to 8 NVIDIA A100 GPUs with a batch size of 16.
We utilized the pre-trained CLIP ViT-B/16 model\footnote{\url{https://github.com/openai/CLIP}} as the backbone VLM for all experiments. We fine-tuned the CLIP image encoder module by employing visual prompt tuning (VPT) approaches~\cite{jia2022visual}, while keeping the CLIP text encoder module frozen.
The number of multiple text prompts was set to $N = 6$ for VOC 2012 and $N = 8$ for PASCAL Context and COCO-Stuff164K datasets.
For the image decoder, we adopted a lightweight transformer consisting of three layers, with the original multi-head attention replaced by our MPSA.
The optimizer was set to AdamW with a specific training schedule for each dataset. 




\begin{table*}[t!]
\caption{Quantitative comparison of zero-shot semantic segmentation performance with baseline methods. The \bf{bold} indicates the best performance.}
\vspace{-2em}
\label{tab_main}
\begin{center}
\resizebox{0.85\linewidth}{!}{
\begin{tabular}{lccccccccc}

\toprule
\multirow{2}{*}{Methods}  & \multicolumn{3}{c}{VOC 2012} & \multicolumn{3}{c}{PASCAL Context} & \multicolumn{3}{c}{COCO-Stuff164K} \\
\cmidrule(lr){2-4} \cmidrule(lr){5-7} \cmidrule(lr){8-10} 
 & mIoU(U) & mIoU(S) & hIoU  & mIoU(U) & mIoU(S) & hIoU  & mIoU(U) & mIoU(S)  & hIoU  \\
\midrule

    \bf{Inductive setting}\\
    \midrule
    ZegFormer~\cite{ding2022zegformer} &  63.6 & 86.4 & 73.3 & - & - & - & 33.2 & 36.6 & 34.8 \\
    Zsseg~\cite{xu2021zsseg} & 72.5 & 83.5 & 77.6 & - & - & - & 36.3 & 39.3 & 37.8 \\
    ZegCLIP~\cite{zhou2023zegclip} & 77.8 & 91.9 & 84.3 & 54.6 & 46.0 & 49.9 & 41.4 & 40.2 & 40.8 \\
    \cmidrule(r){1-1}\cmidrule(r){2-3}	\cmidrule(r){4-4} \cmidrule(r){5-6} \cmidrule(r){7-7} \cmidrule(r){8-9} \cmidrule(r){10-10} 
    {{OTSeg}} & 78.1 & 92.1 & 84.5 & 56.7 & 53.0 & 54.8 & 41.4 & \bf{41.4} & 41.4   \\
     {{OTSeg+}} & \bf{81.6} & \bf{93.3} & \bf{87.1} & \bf{60.4} & \bf{55.2} & \bf{57.7} & \bf{41.8} & {41.3} & \bf{41.5} \\
    \midrule
    \bf{Transductive setting}\\
    \midrule
    Zsseg~\cite{xu2021zsseg} & 78.1  & 79.2 & 79.3 & - & - & - &  43.6  &  39.6  &  41.5  \\
    MaskCLIP+~\cite{zhou2022maskclip} & 88.1 & 86.1 & 87.4 & 66.7 & 48.1 & 53.3 & 54.7& 39.6 & 45.0  \\
    FreeSeg\cite{qin2023freeseg} & 82.6 & {91.8} & 86.9 & - & - & - & 49.1 & {42.2}  & 45.3  \\
    MVP-SEG+\cite{guo2023mvp} & 87.4 & 89.0 & 88.0 & 67.5 & {48.7} & 54.0 & 55.8 &  39.9  & 45.5  \\
    ZegCLIP~\cite{zhou2023zegclip}  & {89.9} & {92.3} & 91.1 & \bf{68.5} & {46.8} & {55.6} & {59.9} & {{40.7}} & {{48.5}}  \\
\cmidrule(r){1-1}\cmidrule(r){2-3}	\cmidrule(r){4-4} \cmidrule(r){5-6} \cmidrule(r){7-7} \cmidrule(r){8-9} \cmidrule(r){10-10} 
    {{OTSeg}} & 94.3  & 94.2  & 94.2 & 66.7  & 53.4 & 59.3 & 60.7 & \bf{41.8}  &  49.5  \\
    {{OTSeg+}} & \bf{94.3} & \bf{94.3}  & \bf{94.4} & {67.0}  & \bf{54.0} & \bf{59.8} & \bf{62.6} & {41.4} & \bf{49.8}   \\

    \midrule
    \bf{Fully-supervised}\\
    \midrule
    ZegCLIP~\cite{zhou2023zegclip}  & {90.9} & {92.4} & 91.6 & \bf{78.7} & {46.5} & {56.9} & {63.2} & {{40.7}} & {{49.6}}  \\
    
    \cmidrule(r){1-1}\cmidrule(r){2-3}	\cmidrule(r){4-4} \cmidrule(r){5-6} \cmidrule(r){7-7} \cmidrule(r){8-9} \cmidrule(r){10-10} 
    
    {{OTSeg}} &94.4&	94.0&	94.2& 78.1 & 55.2 & 64.7& \bf{64.0} & \bf{41.8} & \bf{50.5}  \\
    {{OTSeg+}} & \bf{95.0} & \bf{94.1} & \bf{94.6}  & 78.4 & \bf{54.5} & \bf{65.5} & {63.2} & {41.5} & {50.1}  \\
    
    \bottomrule	
        
\end{tabular}
		}
\end{center}
\vspace{-1.5em}
\end{table*}

\begin{figure*}[!t]
\begin{center}
\includegraphics[width=0.85\linewidth]{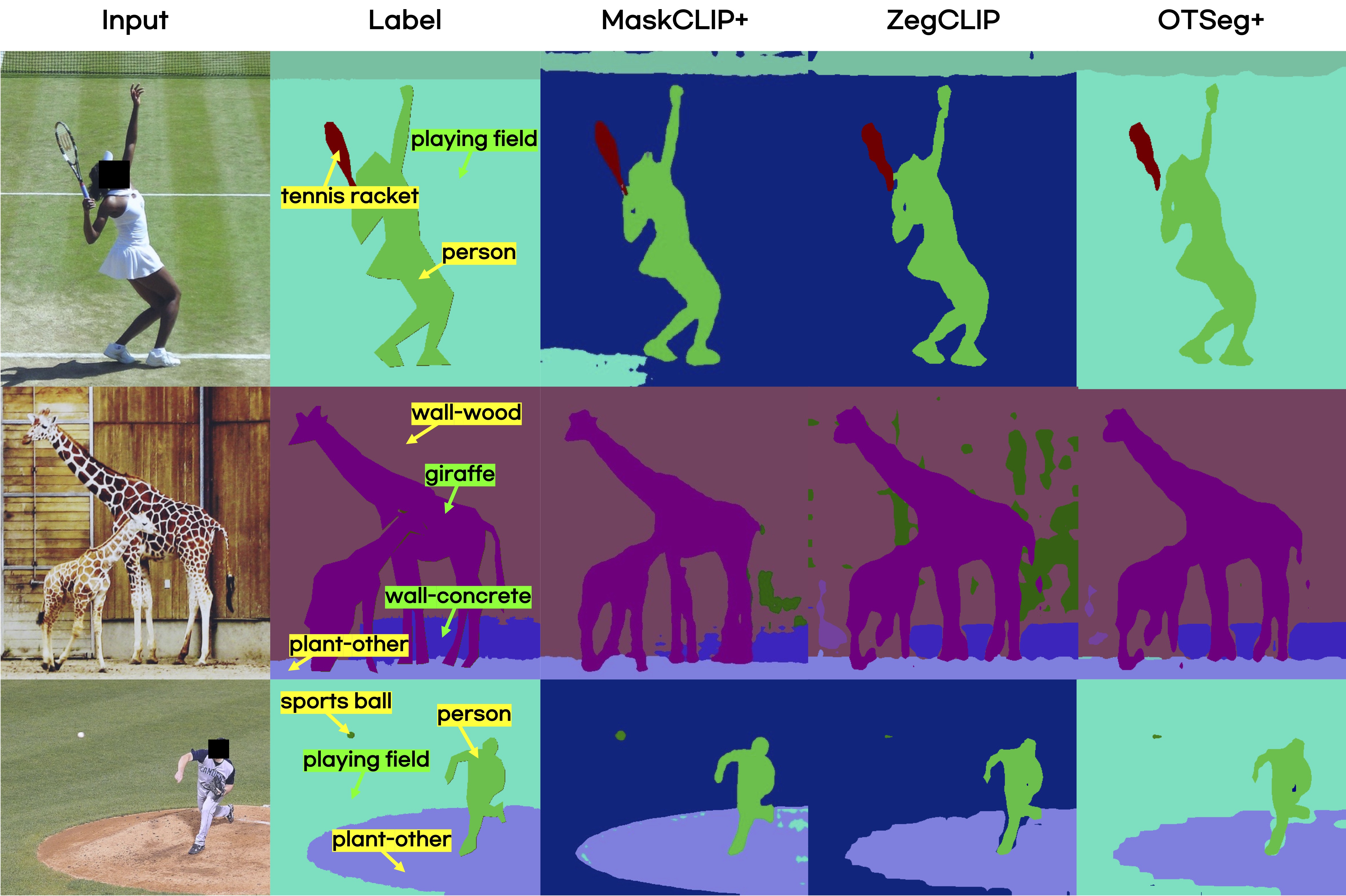}
\caption{Qualitative comparison with previous SOTA models on COCO-Stuff164K dataset. \colorbox{green}{Green} tag indicates unseen classes, while \colorbox{yellow}{yellow} indicates seen classes.} 
\label{fig_seg}
\end{center}
\vspace{-1em}
\end{figure*}

\subsection{Comparison in Zero-Shot Settings}
Quantitative zero-shot segmentation results are presented in \cref{tab_main}. We find that our proposed OTSeg outperforms the previous SOTA in both inductive and transductive settings. It implies that our proposed MPSA module decoder effectively enhances segmentation performance. Furthermore, we observe that our OTSeg+ achieves the best performance on all datasets and demonstrates the effectiveness of our ensemble strategy.
\cref{fig_seg} shows the qualitative zero-shot segmentation performance of our OTSeg+ and other previous approaches.  {OTSeg+ provides the most promising performance for both seen and unseen class objects compared to the previous SOTA methods.} More visual results are provided in \cref{appendix:fig}.

\begin{table}[!t]

\vspace{-1em}
\label{table_gen}
\begin{minipage}[t]{0.5\textwidth}
\caption{{Comparison in cross-data settings.}}
\vspace{-2em}
\begin{center}
    \resizebox{1\linewidth}{!}{
    \begin{tabular}{lccccccc}

    \toprule
    \multirow{3}{*}{Method} & \multirow{3}{*}{Source} &\multicolumn{3}{c}{Target} & \multirow{3}{*}{Source} & \multicolumn{2}{c}{Target} \\
    \cmidrule(lr){3-5} \cmidrule(lr){7-8} 
     &  & \multirow{2}{*}{ADE20K} & \multirow{2}{*}{\shortstack[c]{{PASCAL}\\{Context}}} & \multirow{2}{*}{\shortstack[c]{{VOC}\\{2012}}} & & \multirow{2}{*}{\shortstack[c]{{COCO-}\\{Stuff164K}}} & \multirow{2}{*}{\shortstack[c]{{VOC}\\{2012}}} \\ 
     \\
    \cmidrule(r){1-1} \cmidrule(l){2-2} \cmidrule(l){3-5} \cmidrule(l){6-6}\cmidrule(l){7-8} 
    \multicolumn{7}{l}{\bf{Inductive setting}}\\
    \midrule
     ZegFormer & \multirow{5}{*}{\shortstack[c]{COCO-156}} & 16.4 & - & 80.7  & \multirow{5}{*}{\shortstack[c]{Context-49}} & - & - \\
     Zsseg &  & 15.3 & - & 74.5  &  & - & - \\
     ZegCLIP &  & 19.0 & 41.2 & 93.4 & & 15.6 & 84.0 \\
     \cmidrule(r){1-1} \cmidrule(l){3-5} \cmidrule(l){7-8} 
     OTSeg &  & \bf{20.5} &\bf{49.3} & \bf{94.1} & & 16.7 & 82.7 \\
     OTSeg+ &  & 19.6 &{49.1} & \bf{94.1} & & \bf{16.8} & \bf{85.2} \\

     \midrule
     \multicolumn{7}{l}{\bf{Transductive setting}}\\
     \midrule
      ZegCLIP & \multirow{3}{*}{\shortstack[c]{COCO-156}} & 21.1 &45.8 & 94.2  & \multirow{3}{*}{\shortstack[c]{Context-49}} & {18.1} & 90.6 \\
      OTSeg &  & \bf{21.9} &52.9 & 94.2  &  & \bf{18.9} & 92.2 \\
      OTSeg+ &  & 21.1 & \bf{53.4}  &  \bf{94.4} & & {18.1} & \bf{92.4} \\
    
    \bottomrule
\end{tabular}
}
\end{center}
\label{tbl:cross}
\end{minipage}\hfill
\begin{minipage}[t]{0.45\textwidth}
	\caption{{Comparison of memory cost and inference time. All models are evlauated on a single 3090 GPU.}}
	\vspace{-2em}
	\begin{center}
		\resizebox{1\linewidth}{!}{
		\begin{tabular}{lccc} 
			\toprule				
			Method & $\#$ Parameter (M) $\downarrow$ & GFLOPS  $\downarrow$ & FPS $\uparrow$  \\				
			\cmidrule(l){1-1} \cmidrule(lr){2-2} \cmidrule(lr){3-3} \cmidrule(lr){4-4} 
			Zsseg & 61.1 & 1916.7 & 4.2 \\
			ZegFormer & 60.3 & 1829.3 & 6.8 \\
			ZegCLIP & \bf{13.8} & \bf{61.1} & \bf{25.6} \\
			\cmidrule(l){1-1} \cmidrule(lr){2-2} \cmidrule(lr){3-3} \cmidrule(lr){4-4} 
			OTSeg & \bf{13.8} & ${61.9}_{\textcolor{red}{-0.8}}$ & ${23.6}_{\textcolor{red}{-2.0}}$  \\
			OTSeg+ & \bf{13.8} & ${61.9}_{\textcolor{red}{-0.8}}$ & ${22.5}_{\textcolor{red}{-3.1}}$ \\
			\bottomrule				
		\end{tabular}
	}
	\end{center}
	\label{tbl:efficieny}
\end{minipage}
\vspace{-2em}
\end{table}

\subsection{Comparison in Cross-Dataset Settings}
To further evaluate the generalization capabilities of OTSeg, we perform cross-dataset experiments across COCO-Stuff164K and PASCAL Context datasets. {The model is exclusively trained on the source dataset with labels for seen classes, as indicated as COCO-156 and Context-49, respectively, and then evaluated on the target dataset without any fine-tuning. As for the evaluation target, we added a challenging ADE20K~\cite{zhou2019semantic} dataset.} We compare the results in both inductive and transductive settings. As shown in \cref{tbl:cross}, Our proposed methods, both OTSeg and OTSeg+, outperform the previous SOTA methods and demonstrate superior generalization performance in both experimental settings.

\subsection{Comparison on Efficiency}
To validate the efficiency of OTSeg, we compare the number of learnable parameters, training complexity (GFLOPS), and inference speed with other baseline approaches in \cref{tbl:efficieny}. Our OTSeg does not require additional learnable parameters compared to ZegCLIP, yet achieves the best performance while {sacrificing slightly increased GFLOPS (1.2$\%$) and decreased} FPS (7.8$\%$). Note that our method is still 4-5 times faster than other {mask proposal-based} two-stage methods such as Zsseg and ZegFormer.

\begin{figure*}[!t]
\begin{center}
\includegraphics[width=0.9\linewidth]{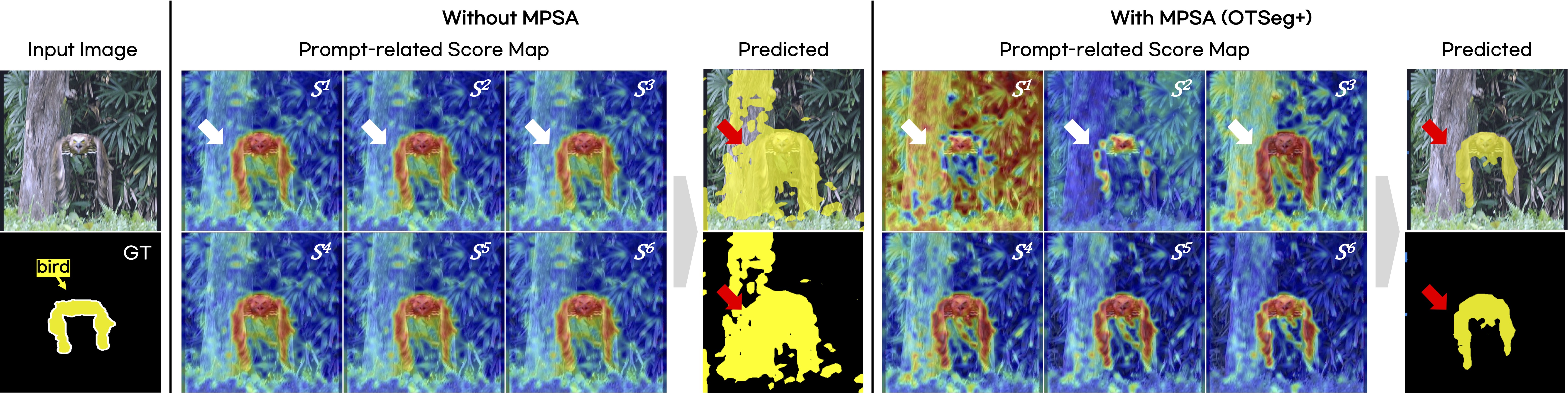}
\caption{Visual comparison of prompt-related score map. While all the text prompt-related score map ${S}^i$ are cohered without MPSA, with our MPOT,
each ${S}^i$ is diversely activated and focuses on different semantic
attributes (white arrows), which helps the model effectively differentiates the target object from the background (red arrows).} 
\label{fig_score}
\end{center}
\vskip -0.1in
\vspace{-1.5em}
\end{figure*}

\subsection{Ablation Studies}
\textbf{MPSA for Enhanced Multimodal Alignment}
In \cref{fig_intro}, we visualize the effectiveness of MPSA by showing each text prompt-related score map, which focuses on different semantic features. We further analyze the reason behind the contribution of MPSA to produce better segmentation results by comparing each text prompt-related score map {and their corresponding segmentation results} in \cref{fig_score}. With MPSA, we observe that each prompt-related score map is diversely dispersed, whereas the baseline method without MPSA shows significantly cohered score maps, as indicated in white arrows. {This visual result suggests that MPSA is helpful in differentiating various semantic attributes from the target object, which yields the final score map selectively focuses on target-related features, leading improved performance on classnames-driven semantic segmentation.\\}

\noindent\textbf{{Analysis on OTSeg Ensemble Component}}
{To investigate the effectiveness of our ensemble {strategy which combines the decoder output and the refined score map, we ablate each path prediction from the ensembled prediction, as shown in \cref{ensemble}.} We provide the results for both the inductive and the transductive settings, as well as the fully supervised setting to establish the upper bound of each approach. Remarkably, {depending solely on the decoder output or the refined score map}, which corresponds to the effect of MPS or MPSA, demonstrates superior performance in all the setting. This suggests that the proposed MPS or MPSA effectively enhances zero-shot semantic segmentation performance. {Despite each path prediction itself outperforms previous SOTA approaches}, the proposed ensemble strategy yields the best performance in almost settings, validating the rationale of our ensemble approach.}

\begin{table*}[!t]
\caption{Analysis on OTSeg Ensemble Components.}
\vspace{-2em}
\label{tab:ablation}
\begin{center}
\resizebox{0.85\linewidth}{!}{
\begin{tabular}{ccccccccccc}

\toprule
\multicolumn{2}{l}{Model Predictions} &   \multicolumn{3}{c}{VOC 2012} & \multicolumn{3}{c}{PASCAL Context} & \multicolumn{3}{c}{COCO-Stuff164K} \\
\cmidrule(r){1-2} \cmidrule(r){3-5} \cmidrule(r){6-8} \cmidrule(r){9-11} 
Decoder & ScoreMap & mIoU(U) &  mIoU(S) & hIoU  & mIoU(U) &  mIoU(S) & hIoU  & mIoU(U) &  mIoU(S) & hIoU \\
\midrule
\multicolumn{2}{l}{\bf{Inductive setting}}\\
\midrule
\cmark &  \xmark & 78.1 & 92.1 & 84.5  & 56.7 & 53.0 & 54.8 & 41.4 & \bf41.4& 41.4\\
\xmark &  \cmark & 73.7 & 92.7 & 81.8  & 55.6 & 54.3 & 54.9 & 37.9 & 40.7 & 39.3\\

\cmidrule(r){1-1} \cmidrule(r){2-2} \cmidrule(r){3-4} \cmidrule(r){5-5} \cmidrule(r){6-7}  \cmidrule(r){8-8} \cmidrule(r){8-8} \cmidrule(r){9-10}  \cmidrule(r){11-11} 

\cmark &  \cmark & \bf{81.6} &\bf{93.3} & \bf{87.1}  & \bf{60.4} & \bf{55.2} & \bf{57.7} & \bf{41.8} & {41.3} & \bf{41.5} \\
\midrule
\multicolumn{2}{l}{\bf{Transductive setting}}\\
\midrule
\cmark &  \xmark & 94.3 & 94.2 & {94.2}   & 66.7  & 53.4 & 59.3 & 60.7 & \bf{41.8} & 49.5  \\
\xmark &  \cmark & \bf{94.4} & \bf{94.3} & \bf{94.3}   & 66.7  & 53.4 & 59.3 & 58.9 & 40.9 & 48.3   \\

\cmidrule(r){1-1} \cmidrule(r){2-2} \cmidrule(r){3-4} \cmidrule(r){5-5} \cmidrule(r){6-7}  \cmidrule(r){8-8} \cmidrule(r){9-10}  \cmidrule(r){11-11} 

\cmark &  \cmark & 94.3 & \bf{94.3} & \bf{94.3}   & \bf{67.0}  & \bf{54.0} & \bf{59.8} & \bf{62.6} & {41.4} & \bf{49.8}  \\
\midrule
\multicolumn{2}{l}{\bf{Fully-supervised}}\\
\midrule
\cmark &  \xmark & 94.4 & 94.0 & {94.2} & 78.1  & 55.2  & 64.7 & \bf64.0 & \bf41.8 & \bf50.5  \\
\xmark &  \cmark & {94.7} & \bf{94.3} & {94.5}   & 77.8 & 55.4 & 64.7 & {62.5} & {40.9} & {49.5}  \\

\cmidrule(r){1-1} \cmidrule(r){2-2} \cmidrule(r){3-4} \cmidrule(r){5-5} \cmidrule(r){6-7}  \cmidrule(r){8-8} \cmidrule(r){9-10}  \cmidrule(r){11-11}

\cmark &  \cmark & \bf{95.0} & {94.1} & \bf{94.6}   & \bf{78.4}  & \bf{56.2} & \bf{65.5} & {63.2} & {41.5} & {50.1} \\

\bottomrule

\end{tabular}
}\label{ensemble}
\end{center}
\vspace{-1.5em}
\end{table*}

\begin{table*}[t!]
\caption{Component analysis of MPSA module under the inductive setting.}
\vspace{-2em}
\label{tab:ablation}
\begin{center}
\resizebox{0.85\linewidth}{!}{
\begin{tabular}{llccccccccccc}

\toprule
\multirow{2}{*}{}  & \multirow{2}{*}{Component} &  \multirow{2}{*}{Multi Text Prompt} & \multirow{2}{*}{Configuration} &  \multicolumn{3}{c}{VOC 2012} & \multicolumn{3}{c}{PASCAL Context} \\
\cmidrule(lr){5-7} \cmidrule(lr){8-10} 

& ~ & ~ & & mIoU(U) &  mIoU(S) & hIoU  & mIoU(U) &  mIoU(S) & hIoU  \\

\cmidrule(r){2-2} \cmidrule(r){3-3} \cmidrule(r){4-4} \cmidrule(r){5-6}  \cmidrule(r){7-7} \cmidrule(r){8-9} \cmidrule(r){10-10}
 & \multirow{3}{*}{(a) Number of Prompt}& \multirow{3}{*}{\cmark} &4 & 76.5 & 93.3  & 84.1 & 44.8 & 52.0 & 48.1 \\
& & & 6 & \textbf{81.6} & \textbf{93.3}  & \textbf{87.1}  &  {48.1} & {52.8} & {50.3} \\
& & & 8 & 79.9 & 93.3 & 86.1 & \bf{60.4} & \bf{55.2} & \bf{57.7} \\

\cmidrule(r){2-2} \cmidrule(r){3-3} \cmidrule(r){4-4} \cmidrule(r){5-6}  \cmidrule(r){7-7} \cmidrule(r){8-9} \cmidrule(r){10-10}
  & \multirow{3}{*}{(b) Matching Method} 
& \xmark & \multirow{2}{*}{Cross-attention} & 77.9 & 91.8 & 84.3 & 50.9 & 49.7 & 50.3 \\ %
& & \cmark &  & 67.1 & 92.1 & 77.6 & 57.2 & 52.0 & 54.5 \\ 
\cmidrule(r){3-3} \cmidrule(r){4-4} \cmidrule(r){5-6}  \cmidrule(r){7-7} \cmidrule(r){8-9} \cmidrule(r){10-10}
 & & \cmark &  MPSA & \bf{81.6} & \bf{93.3} & \bf{87.1}   & \bf{60.4}  & \bf{55.2} & \bf{57.7} \\

 
\bottomrule

\end{tabular}
}
\end{center}
\vspace{-2em}
\end{table*}

{\subsection{Component Analysis}
To study effect of the component of OTSeg on the zero-shot segmentation performance, we conduct a component analysis in \cref{tab:ablation}, which includes: the number of text prompts and the matching method.\\}

\noindent\textbf{{Number of Text Prompts}} In \cref{tab:ablation}(a), we observe segmentation performance by varying the total number $N$ of the introduced multiple text prompts. Our empirical findings indicate that OTSeg achieves optimal performance when $N = 6$ for VOC 2012 dataset, and $N = 8$ for PASCAL Context dataset. {This suggests that, while $N = 6$ is sufficient for datasets with fewer classes, a way increased number of text prompts can add additional performance gains, particularly for larger datasets such like PASCAL Context and COCO-Stuff164K, which may assist the model in acquiring varied semantic features related to text prompts.\\}

\noindent\textbf{{Matching Method}} {In \cref{tab:ablation}(b), we further compare our MPSA matching method with the conventional cross-attention and its variants.}
We observe that the naive extension of multiple text prompts for the cross-attention mechanism results in detrimental effects on specific datasets, such as VOC 2012. Whereas, our MPSA demonstrates its superior performance with margins of 3$\%$ and 7$\%$ hIoU compared to the cross-attention method. These results demonstrate the reason how OTSeg achieves the best performance in zero-shot segmentation settings, which is not only rooted from leveraging multiple text prompts, but also from the proposed Sinkhorn matching mechanism, which optimally transports multiple text prompts to related pixel embeddings.








\section{Discussion and Limitation}
In this study, we demonstrate through MPS and MPSA that each text prompt serves as a score map capable of capturing different semantic features. However, our  each score map is limited by the fact that the true meanings of text prompts are not fully captured. This issue arises from the framework that does not consider the association between semantic meanings and visual features. {Furthermore, even though the application of zero-shot semantic segmentation yields noteworthy results with scalability, this module has yet to be applied to other tasks such as open-vocabulary or instance and panoptic segmentation within the scope of our investigation.} These areas will be considered as future research.

\section{Conclusion}
In this study, we introduce OTSeg, a novel multimodal matching framework for zero-shot semantic segmentation, which leverages multiple text prompts with Optimal Transport (OT)-based text-pixel alignment module, specifically Multi-Prompts Sinkhorn (MPS), along with its extension for cross-attention mechanism, Multi-Prompts Sinkhorn Attention (MPSA). By incorporating MPSA within Transformer, our proposed OTSeg effectively aligns semantic features between multiple text prompts and image pixels, {and selectively focuses on target object-related features.} Through extensive experiments, we demonstrate OTSeg's capability to achieve the state-of-the-art (SOTA) performance on zero-shot segmentation (ZS3) tasks across three benchmark datasets. We believe that OTSeg {can contribute in opening new directions for future researches in multimodal alignment and zero-shot learning, with potential applications in various domains requiring multi-conceptual semantic understandings of vision.}
{\\}

\noindent\textbf{Acknowledgments}
This work was supported by Institute for Information $\&$
communications Technology Promotion(IITP) grant funded by the Korea government(MSIT) (No.RS-2019-II190075) Artificial Intelligence Graduate School Program(KAIST), and also supported by the National Research Foundation of Korea(NRF) grant funded by the Korea government(MSIT) (RS-2024-00345854), (RS-2024-00336454), (RS-2023-00262527), and also supported by Field-oriented Technology Development Project for Customs Administration funded by the Korea government through the National Research Foundation (NRF) of Korea under Grant NRF2021M3I1A1097910, and also Administration by Culture, Sports and Tourism R$\&$D Program through the Korea Creative Content Agency grant funded by the Ministry of Culture, Sports and Tourism in 2023.
 \\

\bibliographystyle{splncs04}
\bibliography{egbib}



\clearpage
\newpage
\newpage

\appendix
\renewcommand{\thefigure}{S\arabic{figure}}
\renewcommand{\thetable}{S\arabic{table}}
\setcounter{figure}{0}
\setcounter{table}{0}
\section{Algorithm of MPS}\label{sec_algo}

As discussed in~\cref{sec:sinkhorn}, we adopt the log-scaling version of Sinkhorn approaches for our proposed Multiple Prompt Sinkhorn (MPS) algorithm as follow:
 \vspace{-2em}
\begin{algorithm}[!]
	\caption{Multiple Prompt Sinkhorn Algorithm}
	\label{algo-ot}
	\SetKwInOut{Input}{Input}
	\SetKwInOut{Output}{Output}
	\SetKwInput{kwInit}{Initialization}
	\SetKwInput{kwset}{Given}
	\kwset{The feature map size $M = HW$, the number of prompts $N$,  $\bs{\mu} = \bs{1}^{M}/M$ , $\bs{\nu} = \bs{1}^{N}/N$, the score map $\bs{S}$ ;}
	\Input{The cost matrix $\bs{C} = \bs{1} - \bs{S}$, hyper-paramter $\bs{\epsilon}$, the max iteration $t_{\text{max}}$;}
	\kwInit{$\bs{K} = \exp(-\bs{C}/\bs{\epsilon})$, $t \leftarrow 1, \bs{b}^0 = 0$;}
	\While{$t \leq t_{\text{max}}$ $\mathbf{\text{and not converge}}$ }{
	$\bs{a}^t = \log{\bs{\mu}} - \log \left[\sum \exp \left[-\frac{1}{\epsilon} \bs{C} + \bs{b}^{t-1})\right] \right]$; \\ 
	$\bs{b}^t = \log{\bs{\nu}} - \log \left[\sum \exp \left[-\frac{1}{\epsilon} \bs{C}^{\top} + \bs{a}^{t-1})\right] \right]$; } 
	\Output{Optimal transport plan $\bs{T}^{\ast}$ = $\text{diag}(\exp(\bs{a}))^t\bs{K}\text{diag}(\exp(\bs{b}))^t$ ;}
\end{algorithm}

\section{Details of Loss function}
\label{appendix:loss}
As discussed in ~\cref{sec:loss}, we combine three different losses, including the focal loss based on Cross Entropy (CE) loss, and the dice loss, which are given  by:
\begin{align}
\begin{split}
\mathcal{L}_{\text{ce}} = -\frac{1}{HW} \sum^{HW}_{i=1} \gty_i \log(\phi(\mbf{Y}_i)) \\
    + (1- \gty_i)\log(1-\phi(\mbf{Y}_i)), \\   
\end{split}\\
\begin{split}
\mathcal{L}_{\text{focal}} = -\frac{1}{HW} \sum^{HW}_{i=1}\gty_i (1- \sigma(\mbf{Y}_i))^{\gamma} \log(\sigma(\mbf{Y}_i) )\\
    + (1- \gty_i)\sigma(\mbf{Y}_i)^{\gamma}\log(1-\sigma(\mbf{Y}_i)), \\   
\end{split}\\    
\begin{split}
\mathcal{L}_{\text{dice}} = 1 -\frac{2\sum^{HW}_{i=1}\gty_i \mbf{Y}_i}{\sum^{HW}_{i=1} {\gty_i}^2 + \sum^{HW}_{i=1} {\mbf{Y}_i}^2}, 
\end{split} 
\end{align}
where $\mbf{Y}$ is the model decoder outputs, $\gty$ is the ground truth label, 
$\sigma(\cdot)$ is Sigmoid operations, 
$\gamma$ is a hyper-parameter to balance hard and easy samples, which is set to 2.
Throughout the entire experiments,
$\lambda_{\text{ce}}$, $\lambda_{\text{focal}}$ and $\lambda_{\text{dice}}$ are set to 1, 20, and 1, respectively.

\section{Relation Descriptor}
\label{appendix:RD}
As discussed in \cref{task}, when an image is inputted into the CLIP image encoder, we obtain the ([CLS] token) $\mathbf{\bar h}_{\text{img}} \in \mathbb{R}^{1 \times D}$ and the pixel embedding $\hii$ in the last layer. To merge these CLIP's text and pixel embeddings, we utilize the relationship descriptor following an approach introduced in~\cite{zhou2023zegclip} to yield the refined text embedding $\hht$ as follows:
$$ \htt = \mathcal{R}_{\psi} (\text{cat}\left[\hii \odot \hht, \hht \right] )  \in \mathbb{R}^{KN \times D}$$
where $\mathcal{R}_{\psi}$ denotes a linear layer for matching the concatenated embedding dimension to the original dimension $D$, and \text{cat} is the concatenation operator, $\odot$ is the Hadamard product.

\section{Definition of Harmonic mean IoU}
\label{appendix:hIoU}
Following the previous works ~\cite{xu2021zsseg,zhou2022maskclip,zhou2023zegclip}, we define harmonic mean IoU (hIoU) among the seen (S) and unseen (U) classes as:
\begin{align}
    \text{hIoU} =\frac{2 * \text{mIoU (S) * mIoU (U)}}{\text{mIoU (S) + mIoU (U)}}
\end{align}

\section{Details of Dataset}
\label{appendix:dataset}

We utilize a total of three datasets, $\textit{i.e.,}$ VOC 2012~\cite{everingham2012pascal}, PASCAL  Context~\cite{mottaghi2014role}, and COCO-Stuff164K~\cite{caesar2018coco}. 
We divide seen and unseen classes for each dataset, following the settings of previous methods~\cite{bucher2019zero,xu2021zsseg,ding2022zegformer,zhou2022maskclip,zhou2023zegclip}. VOC 2012 consists of 10,582 / 1,449 images with 20 categories, for training / validation. The dataset is divided into 15 seen and 5 unseen classes. PASCAL Context is an extensive dataset of PASCAL VOC 2010 that contains 4,996 / 5,104 images for training / test with 60 categories. The dataset is categorized into 50 seen and 10 unseen classes. COCO-Stuff 164K is a large-scale dataset that consists of 118,287 / 5,000 images for training / validation with 171 classes. The dataset is categorized into 156 seen and 15 unseen classes.

\section{Implementation Detail}
\label{appendix:implement}
 We further declare the implementation detail for our work. 
 Input image resolution is set as 480$\times$480 for PASCAL Context, and 512$\times$512 for the rest of the datasets. For training, we choose the lightest training schedule. For the inductive settings, the total training iterations were 20K for VOC 2012, 40K for PASCAL Context, and 80K for COCO-Stuff164K. For the transductive settings, the model was trained on seen classes in the first half of training iterations and then applied self-training by generating pseudo-labels in the remaining iterations, following previous approaches~\cite{zhou2022maskclip,zhou2023zegclip}. The balance factor $\lambda$ for the score map in~\cref{eq:predictfinal} is set to 0.5.

\section{Further Comparison Studies}
\label{appendix:compare}

{To further validate the generalizability of our proposed method in the open-vocabulary segmentation (OVS) setting, we compared our OTSeg and OTSeg+ with FreeSeg \cite{qin2023freeseg} and SAN \cite{xu2023side}. It's noteworthy that OTSeg is designed for ZS3, utilizing a partial dataset of 156 classes, whereas FreeSeg and SAN use the entire COCO-Stuff 171 classes dataset tailored for OVS. Despite differing goals and settings, our proposed OTSeg variants demonstrate comparable results in cross-dataset settings, as shown in Table \ref{tab_cross}, validating their effectiveness.}

\begin{table*}[!h]
\caption{Cross-data comparison with open-vocabulary segmentation methods.}

\label{tab_cross}
\begin{center}
\resizebox{0.9\linewidth}{!}{
\begin{tabular}{llcccc}
\toprule
Category & Method & Source & ADE20K & PC59 & VOC \\
\cmidrule(lr){1-1}  \cmidrule(lr){2-2} \cmidrule(lr){3-3}  \cmidrule(lr){4-6}
\multirow{2}{*}{Zero-shot segmentation (ZS3)} & OTSeg & \multirow{2}{*}{COCO-156} & 21.9 & 52.9 & 94.2 \\
 & OTSeg+ &  & 21.1 & 53.4  & {94.4} \\
\cmidrule(lr){1-1}  \cmidrule(lr){2-2} \cmidrule(lr){3-3}  \cmidrule(lr){4-6}
\multirow{2}{*}{Open-vocabulary segmentation (OVS)} & Freeseg \cite{qin2023freeseg} & \multirow{2}{*}{COCO-171} & 17.9 & 34.4 & 85.6 \\
 & SAN\cite{xu2023side} &  & {27.5} & 53.8 & 94.1 \\
\bottomrule
\end{tabular}
}
\end{center}
\end{table*}

\vspace{-2em}


\section{Robustness of OTSeg with Multiple Seeds}
\label{appendix:compare}

{To evaluate the robustness of our proposed Multiple Prompt Sinkhorn (MPS) algorithm, we repeatedly trained both OTSeg and OTSeg+  with multiple seeds and compared the results on the VOC dataset. Table \ref{table_seed} confirms the robustness and consistent performance of our proposed method across various seeds.}

\begin{table*}[!h]
\caption{Quantitative results with multiple seeds on VOC dataset and comparison in cross-data settings with further approaches.}

\label{table_seed}
\begin{center}
\resizebox{0.7\linewidth}{!}{
\begin{tabular}{clcccccc}
\toprule
\multirow{2}{*}{Settings} & \multirow{2}{*}{Method} & \multicolumn{4}{c}{Seed} & \multirow{2}{*}{Mean $\pm$ STD} \\
\cmidrule(lr){3-6} 
 &  & \#1 & \#2 & \#3 & \#4  &  \\
\cmidrule(lr){1-1} \cmidrule(lr){2-2} \cmidrule(lr){3-6} \cmidrule(lr){7-7}
\multirow{2}{*}{Inductive} & OTSeg & 84.5 & 84.9 & 84.3 & 84.0 & 84.4 $\pm$ 0.4 \\
 & OTSeg+ & 87.1 & 88.0 & 88.0 & 86.5 & 87.4 $\pm$ 0.7 \\
\cmidrule(lr){1-1} \cmidrule(lr){2-2} \cmidrule(lr){3-6} \cmidrule(lr){7-7}
\multirow{2}{*}{Transductive} & OTSeg & 94.2 & 94.6 & 94.2 & 94.3 & 94.4 $\pm$ 0.2 \\
 & OTSeg+ &  94.3 & 94.5 & 94.2 & 94.3 & 94.3 $\pm$ 0.1 \\
\bottomrule
\end{tabular}
}
\end{center}
\end{table*}
\vspace{-1em}


\section{Further Component Analysis}
In this section, we conduct additional component analysis by varying the conditions comprising our proposed method.

\noindent\textbf{Balance Factor} In OTSeg+, we generate the final prediction by ensembling the decoder path and score map path as indicated in \cref{eq:predictfinal}. \cref{tab:ablation_sup} (a) represents the analysis of the balance factor $\lambda$. In inductive settings, the effect of the balance factor is significant compared to transductive settings. We empirically determine our default configuration of the balance factor as 0.5.

\noindent\textbf{Learnable Module} 
To validate the rationale behind our choice of the learnable module, we conduct experiments by altering the conditions outlined in \cref{tab:ablation_sup} (b). We observe that training both the visual prompt in the image encoder and the text prompt in the text module results in a significant drop in performance due to overfitting. While tuning only the learnable text prompt provide relatively improved performance, we observe that our adoption of only tuning the image encoder via visual prompt tuning leads to the best performance.

\begin{table*}[!h]

\caption{Further component analysis on the VOC 2012 dataset.}
\vspace{-1em}
\label{tab:ablation_sup}
\begin{center}
\resizebox{1\linewidth}{!}{
\begin{tabular}{llccccccccc}

\toprule
\multirow{2}{*}{}  & \multirow{2}{*}{Component} &  \multirow{2}{*}{   Ours   } & \multirow{2}{*}{Factor $\lambda$ } & \multirow{2}{*}{Module} & \multicolumn{3}{c}{Inductive setting} & \multicolumn{3}{c}{Transductive setting} \\
\cmidrule(r){6-8} \cmidrule(r){9-11} 

& ~ & ~ & & & mIoU(U) &  mIoU(S) & hIoU  & mIoU(U) &  mIoU(S) & hIoU  \\

\cmidrule(r){2-2} \cmidrule(r){3-3} \cmidrule(r){4-4} \cmidrule(r){5-5}  \cmidrule(r){6-7} \cmidrule(r){8-8} \cmidrule(r){9-10} \cmidrule(r){11-11} 
 & \multirow{5}{*}{(a) Balance factor}&  &  0.1 & & 79.6 & 92.9  & 85.7 & 94.2 & 94.2 & 94.2 \\
   &  &  & 0.3 & & 80.4 & 92.8  & 86.2  &  94.2  & \bf{94.3} & {94.2} \\
 &  & \cmark & 0.5 & & \textbf{81.6} & \textbf{93.3}  & \textbf{87.1}  &  \bf{94.4} & \bf{94.3} & \bf{94.3} \\ 
& & & 0.7 & &80.6 & 93.1 & 86.4 & 94.3&94.2 &94.2\\
& & & 0.9 & &78.8& 92.1 & 84.9 & 94.1 & 94.2&94.1\\

\cmidrule(r){2-2} \cmidrule(r){3-3} \cmidrule(r){4-4} \cmidrule(r){5-5}  \cmidrule(r){6-7} \cmidrule(r){8-8} \cmidrule(r){9-10} \cmidrule(r){11-11}
  & \multirow{3}{*}{(b) Learnable Module} 
& \cmark &  & Decoder, Visual Prompts & \bf{81.6} & \bf{93.3} & \bf{87.1}   & \bf{94.4} & \bf{94.3} & \bf{94.3} \\ %
& &  &  & Decoder, Visual Prompts, Text Prompts & 35.9  & 85.6 & 50.7 & 48.6 & 94.1 & 64.1 \\ 
 & &  &  &  Decoder, Text Prompts  & 42.9 &  89.8 & 58.1 & 45.6 & 91.9 & 61.0 \\
\bottomrule

\end{tabular}
}
\end{center}
\end{table*}


\section{Analysis on Sinkhorn Convergence Comparison Studies}
\label{appendix:converge}

\begin{wrapfigure}{r}{0.48\textwidth}
\vspace{-2.5em}
    \begin{center}
    \includegraphics[width=0.45\textwidth]{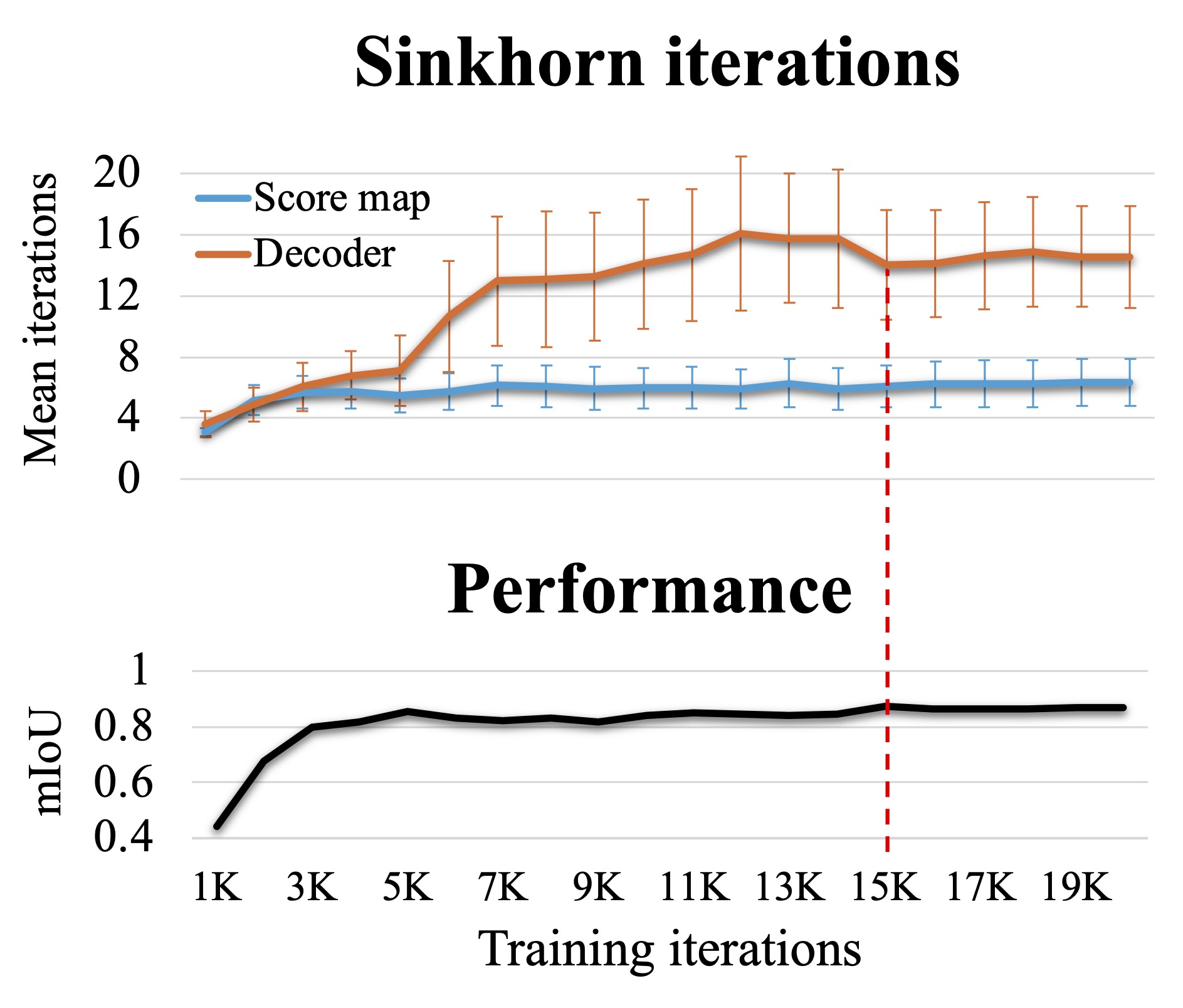}
    \caption{Sinkhorn convergence.}
    \label{fig_convergence}
    \end{center}
\end{wrapfigure}

We further analyzed the convergence of the Sinkhorn algorithm and its effect on performance. We established a defined threshold and applied the Sinkhorn algorithm iteratively until the total errors dropped below this threshold. In Figure~\ref{fig_convergence}, we report that the Sinkhorn algorithm converged in 6 iterations for the score map and 14 iterations for the OT-adapted decoder part on the VOC test dataset. Additionally, we observed that the convergence of the Sinkhorn algorithm around 15K training iterations led to the best performance.

\clearpage



\section{Additional Visual Results}
\label{appendix:fig}

We provide additional visual results. 
\cref{fig_seg_coco} show qualitative zero-shot segmentation performance for COCO-Stuff164K dataset. We reproduce the segmentation results using publically available weights \cite{zhou2022maskclip, zhou2023zegclip}. Our OTSeg demonstrates superior performance in  precisely sectioning boundaries of both seen and unseen objects, unlike previous SOTA method which fail to classify the category or produce noisy results. 





\begin{figure*}[!h]
\vspace{-1em}
\begin{center}
\includegraphics[width=0.97\linewidth]{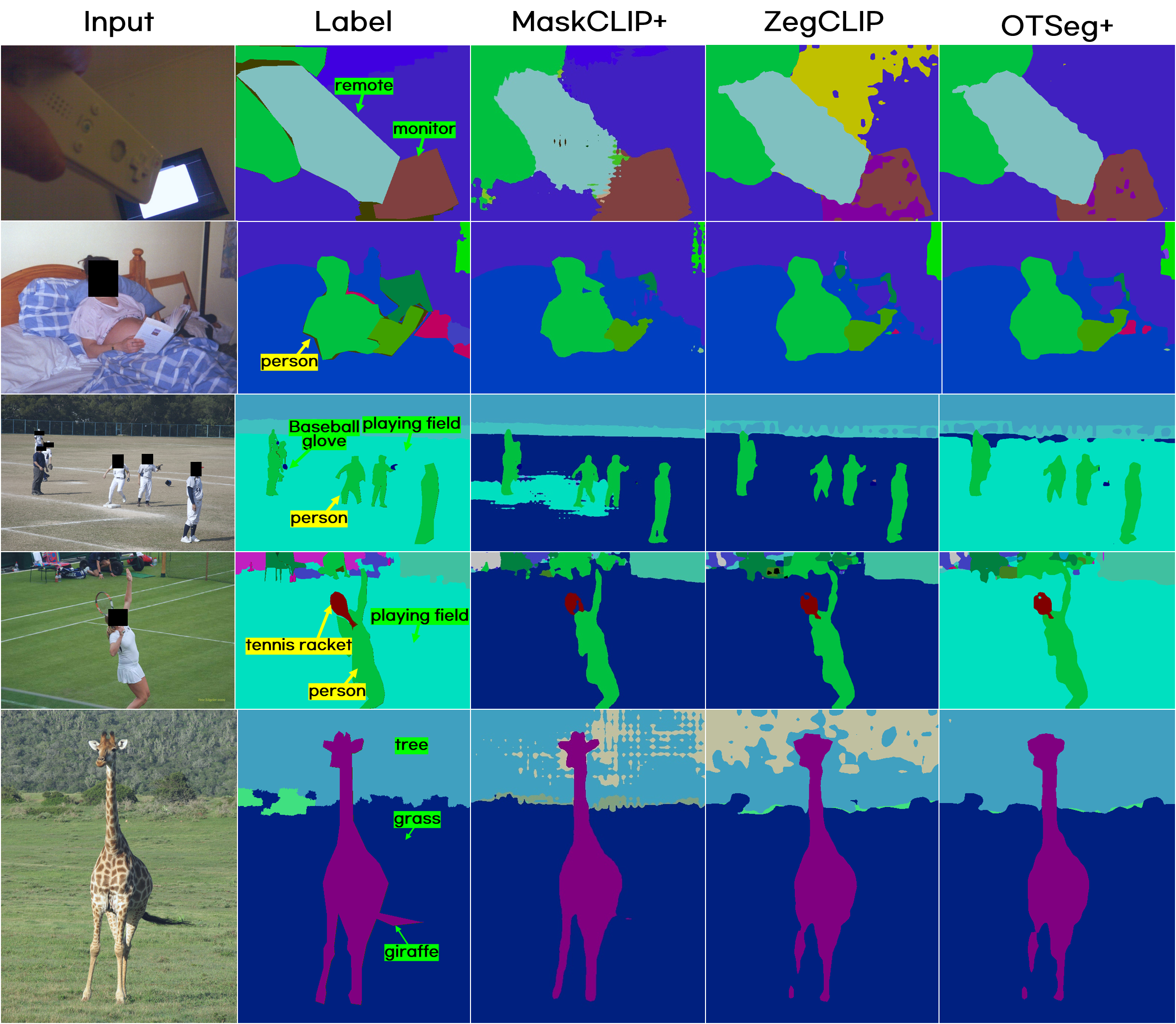}
\caption{Qualitative zero-shot segmentation results of COCO-Stuff164K dataset. The  \colorbox{yellow}{yellow} tag indicates seen classes, while the \colorbox{green}{green} tag indicates unseen classes. }
\label{fig_seg_coco}
\end{center}
\end{figure*}
\vspace{-2em}


\section{Further Score Map Comparison with or without MPSA}
We provide additional visual comparison of prompt-related score map with or without MPSA in \cref{fig_score_sup}. While all the text prompt-related score map ${S}^i$ are cohered without MPSA (white arrows), with our MPOT,
each ${S}^i$ focuses on different semantic
attributes (red arrows), which helps the model utilizes various score maps to differentiate each class name-driven object boundary.

\begin{figure*}[!h]
\begin{center}
\includegraphics[width=1\linewidth]{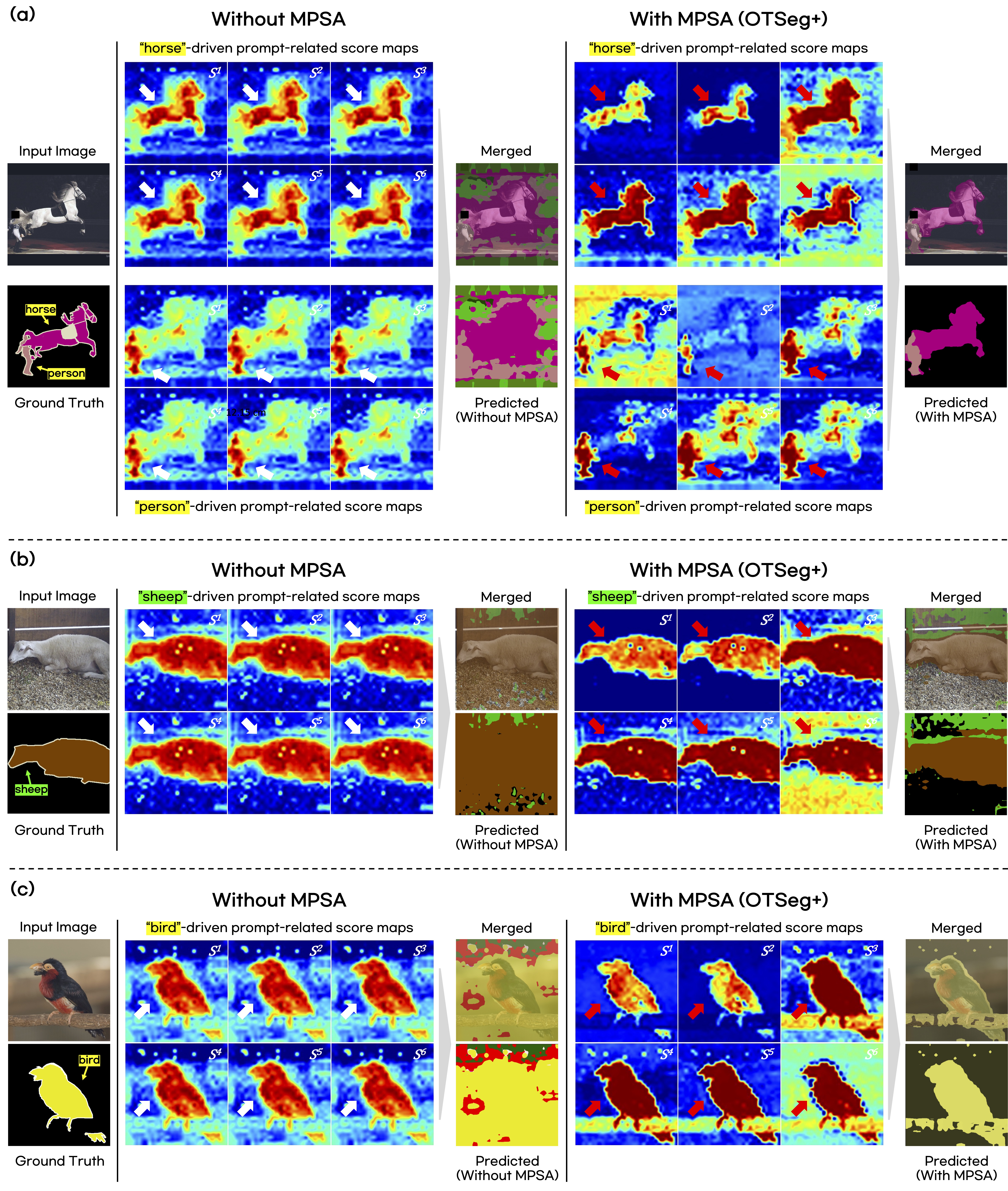}
\caption{Further visual comparison of class name-driven prompt-related score maps. While all the text prompt-related score maps ${S}^i$ are cohered without MPSA (white arrows), with MPSA within our proposed OTSeg+,
each ${S}^i$ is diversely activated (red arrows) to help the model segment each object sharply. The \colorbox{yellow}{yellow} tag indicates seen classes, while the \colorbox{green}{green} tag indicates unseen classes.}
\label{fig_score_sup}
\end{center}
\end{figure*}

\end{document}